\documentclass{article}

\usepackage[preprint,nonatbib]{neurips_2023}

\usepackage[utf8]{inputenc} 
\usepackage[T1]{fontenc}    
\usepackage{hyperref}       
\usepackage{url}            
\usepackage{booktabs}       
\usepackage{amsfonts}       
\usepackage{nicefrac}       
\usepackage{microtype}      

\usepackage{amsmath,amsfonts,bm}

\def\eqref#1{equation~\ref{#1}}

\def\1{\bm{1}}

\DeclareMathAlphabet{\mathsfit}{\encodingdefault}{\sfdefault}{m}{sl}
\SetMathAlphabet{\mathsfit}{bold}{\encodingdefault}{\sfdefault}{bx}{n}

\DeclareMathOperator*{\argmax}{arg\,max}

\usepackage{wrapfig}
\usepackage{enumitem}
\usepackage{graphicx}
\usepackage{comment}
\usepackage{cases}
\usepackage{amsmath,amssymb} 
\usepackage{bbm}
\usepackage{csquotes}

\usepackage{multirow}
\usepackage{booktabs}

\usepackage{pifont}

\usepackage{color, colortbl}
\definecolor{Gray}{gray}{0.9}

\newcommand*{\twoelementtable}[3][l]%
{%
    \begin{tabular}[t]{@{}#1@{}}%
        #2\tabularnewline
        #3%
    \end{tabular}%
}

\usepackage{array}
\usepackage{tabularx}

\usepackage{mathtools}

\usepackage{subfig}
\usepackage{textcomp}
\definecolor{deemph}{gray}{0.6}

\definecolor{baselinecolor}{gray}{.9}

\usepackage[table]{xcolor}

\newlength\savewidth

\title{Prompt Ensemble Self-training for \\
Open-Vocabulary Domain Adaptation}

\author{Jiaxing Huang\thanks{indicates equal contribution.} , Jingyi Zhang\footnotemark[1] ,  Han Qiu, Sheng Jin, Shijian Lu\thanks{corresponding author.}\\
S-lab, School of Computer Science and Engineering, Nanyang Technological University\\
{\tt\small \{Jiaxing.Huang, Jingyi.Zhang, Sheng.Jin, Shijian.Lu\}@ntu.edu.sg} \\
{\tt\small han023@e.ntu.edu.sg}
}

\begin{document}

\maketitle

\begin{abstract}
Traditional domain adaptation assumes the same vocabulary across source and target domains, which often struggles with limited transfer flexibility and efficiency while handling target domains with different vocabularies.
Inspired by recent vision-language models (VLMs) that enable open-vocabulary visual recognition by reasoning on both images and texts, we study open-vocabulary domain adaptation (OVDA), a new unsupervised domain adaptation framework that positions a pre-trained VLM as the source model and transfers it towards arbitrary unlabelled target domains. 
To this end, we design a Prompt Ensemble Self-training (PEST) technique that exploits the synergy between vision and language to mitigate the domain discrepancies in image and text distributions simultaneously. 
Specifically, PEST makes use of the complementary property of multiple prompts within and across vision and language modalities, which enables joint exploitation of vision and language information and effective learning of image-text correspondences in 
the 
unlabelled target domains. Additionally, PEST captures temporal information via temporal prompt ensemble which helps memorize previously learnt target information.
Extensive experiments show that PEST outperforms the state-of-the-art consistently across 10 image recognition tasks.
\end{abstract}

\section{Introduction}
Deep learning-based vision models~\cite{he2016deep,dosovitskiy2020image} have achieved great success in myriad image recognition tasks but at the price of laborious annotation of large-scale training images~\cite{deng2009imagenet}. 
To circumvent the annotation constraint, domain adaptation (DA)~\cite{liang2020we,huang2021model} has been explored to transfer a vision model pre-trained in certain labelled source domains towards unlabelled target domains by mitigating the inter-domain discrepancies in image distributions. 
However, traditional DA~\cite{tzeng2017adversarial,liang2020we,huang2021model} assumes that source and target domains have the same vocabulary. It struggles while handling target domains with different vocabularies, limiting its flexibility and efficiency greatly in unsupervised transfer.

\begin{figure}[t]
\centering
\includegraphics[width=0.98\linewidth]{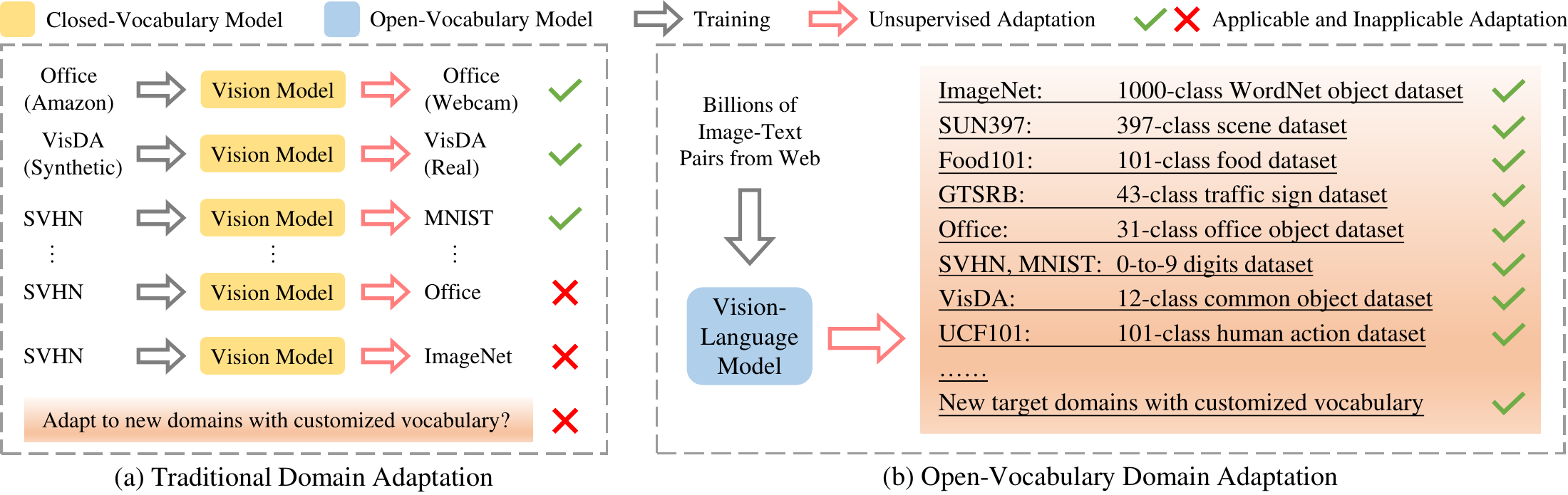}
\caption{
Traditional domain adaptation transfers a vision model across domains of the same vocabulary, which struggles while handling target domains with different vocabularies or new domains with customized vocabularies as illustrated in (a). Inspired by the recent open-vocabulary vision-language models (VLMs), we study open-vocabulary domain adaptation, a new unsupervised domain adaptation framework that positions a single pre-trained VLM as the source model and transfers it towards arbitrary unlabelled target domains as illustrated in (b).
}
\label{fig:introduction}
\vspace{-2ex}
\end{figure}

Inspired by recent vision-language models (VLMs)~\cite{radford2021learning} that enable open-vocabulary visual recognition by reasoning on both images and texts, we study open-vocabulary domain adaptation (OVDA), a new unsupervised domain adaptation (UDA) framework that positions a pre-trained VLM as the source model and transfers it towards arbitrary unlabelled target domains. OVDA requires a single pre-trained VLM only while transferring towards target domains of different vocabularies, instead of preparing multiple vocabulary-specific vision models with respective source datasets, as illustrated in Fig.~\ref{fig:introduction}. 
In addition, OVDA allows unsupervised transfer towards new domains with customized vocabulary, which greatly mitigates the image annotation constraint and facilitates deep network training while handling various new visual recognition tasks. 
On the other hand, the shift from traditional domain adaptation toward OVDA comes with a new challenge, namely, the cross-domain discrepancies in both image distributions and text distributions.

Drawing inspiration from the recent advances in multi-prompt learning~\cite{jiang2020can,schick2020exploiting,qin2021learning,yuan2021bartscore} in natural language processing (NLP), we design PEST, a Prompt Ensemble Self-training technique that exploits the synergy between vision and language to mitigate the domain discrepancies in image and text distributions simultaneously while self-training. PEST makes use of the complementary property of multiple prompts within and across vision and language modalities: it exploits VLMs to encode the image prompts~\cite{luddecke2022image,zang2022open} and text prompts~\cite{luddecke2022image,zang2022open}
into an aligned vision-language feature space and fuses the encoded visual and textual features 
to ``prompt'' unsupervised self-training for more accurate pseudo label prediction. This enables joint exploitation of vision and language information and effective learning of image-text correspondences in the unlabelled target domains. In addition, PEST captures temporal information via temporal prompt ensemble, which helps memorize previously learnt target information by fusing the prompts encoded by the intermediate models evolved along the adaptation process.

The proposed PEST can be viewed as a new type of self-training with multi-prompt learning for the task of OVDA. It has three desirable advantages: 1) it introduces vision prompt ensemble and language prompt ensemble and enables simultaneous mitigation of image and text discrepancies across domains effectively; 2) it introduces temporal prompt ensemble along the adaptation process which allows harvesting previously learnt target information effectively;
3) it works within an aligned image-text feature space which allows prompt ensemble not only within but also across vision, language and temporal dimensions, capturing their complementary advantages effectively.

In summary, the contributions of this work are threefold. \textit{First}, we design a novel open-vocabulary domain adaptation framework that explores multi-prompt learning upon self-training to learn effective image-text correspondences over unlabelled target images. To the best of our knowledge, this is the first work that explores multi-prompt learning for OVDA. \textit{Second}, we design prompt ensemble self-training that introduces prompt ensemble over vision, language and temporal dimensions for simultaneous mitigation of image and text discrepancies in OVDA. \textit{Third}, extensive experiments show that the proposed prompt ensemble self-training outperforms the state-of-the-art consistently across multiple image recognition tasks.

\section{Related Work}

\textbf{Domain Adaptation} (DA), a type of unsupervised transfer learning, aims to adapt a model pre-trained on certain labelled source domains towards unlabelled target domains. Most existing DA methods can be broadly grouped into two categories. The first category employs \textit{adversarial learning} to align source and target image distributions over the input space~\cite{hoffman2018cycada,huang2021fsdr,huang2021rda,zhang2021spectral,sankaranarayanan2018learning, li2019bidirectional, hong2018conditional,yang2020fda}, feature space~\cite{tzeng2017adversarial,tsai2018learning,huang2021mlan,zhang2021detr,wei2021toalign,guan2021domain}, output space~\cite{saito2017adversarial,guan2021uncertainty,luo2019taking,saito2018maximum} or latent space~\cite{vu2019advent,huang2020contextual,tsai2019domain,chen2018road,zhang2017curriculum}, for mitigating the distribution discrepancy across domains. 
The second approach explores \textit{self-training} that learns from unlabelled target images with predicted pseudo labels~\cite{zou2018unsupervised, huang2021model,saleh2018effective,zhong2019invariance,huang2021category,zou2019confidence,huang2021cross,guan2021scale,liang2020we,liang2021source}.
Despite their great success, most existing methods assume the same vocabulary across the source and target domains and cannot handle target domains with different vocabulary or new domains with customized vocabulary. This limits the flexibility and efficiency of DA greatly.
We study open-vocabulary domain adaptation in this work, a new framework that reasons both images and texts and allows unsupervised transfer learning towards arbitrary unlabelled target domains. We design prompt ensemble self-training that explores the synergy of vision and language to mitigate image and text domain gaps simultaneously in OVDA.

\textbf{Vision Language Model} (VLM)~\cite{radford2021learning,jia2021scaling,yuan2021florence,yu2022coca,tschannen2022image,zhang2023vision} aims to learn effective vision-language correlation from image-text pairs that are almost infinitely available on the Web. It has demonstrated great potential in open-vocabulary visual recognition by recognizing images with arbitrary texts.
As a representative, CLIP~\cite{radford2021learning} collects 400 million image-text pairs from the Web and learns rich vision-language correlation via image-text contrastive learning.
Despite its great success, VLMs often suffer from degraded performance due to cross-domain discrepancies with respect to various downstream domains. Unlike recent attempts~\cite{zhou2022learning,zhou2022conditional} that adapt VLMs with few-shot labelled target images, we focus on adapting VLMs towards various downstream domains by ingeniously exploiting the unlabelled target images which are often off-the-shelf available in abundance.

\textbf{Multi-Prompt Learning} explores complementary advantages of different prompts~\cite{jiang2020can} which was originally designed for effective transfer of large language models in NLP. Most existing methods can be broadly grouped into three categories. The first is \textit{prompt ensembling} that creates multiple unanswered prompts for an input to predict via uniform averaging~\cite{jiang2020can,schick2020exploiting,yuan2021bartscore}, weighted averaging~\cite{jiang2020can,qin2021learning,schick2020exploiting}, majority voting~\cite{lester2021power,hambardzumyan2021warp}, etc. The second exploits \textit{prompt augmentation} that provides several answered prompts for an input for better predictions, where most studies focus on the selection~\cite{gao2020making,lu2021fantastically,liu2021makes} and ordering~\cite{lu2021fantastically,kumar2021reordering,guu2018generating} of answered prompts. The third works by \textit{prompt composition or decomposition}~\cite{han2022ptr,cui2021template}, which constructs multiple sub-prompts for better predictions.

\section{Method}

\subsection{Preliminaries of Vision-Language Model}

\textbf{Vision-language model (VLM) training.} VLM~\cite{radford2021learning,jia2021scaling,yuan2021florence,yu2022coca,tschannen2022image} learns effective vision-language correlation from image-text pairs that are almost infinitely available on the Web~\cite{radford2021learning,schuhmann2021laion}. The training involves a VLM $F = \{F^{I}, F^{T}\}$ where $F^{I}$ and $F^{T}$ denote an image encoder and a text encoder respectively, and an image-text dataset $D_{s} = \{(x^{I}_{n}, x^{T}_{n})\}_{n=1}^{N}$ where $x^{I}_{n}$ and $ x^{T}_{n}$ stand for an image sample and its paired text sample. Given $F$ and $D_{s}$, rich vision-language correlation can be learnt with a vision-language training objective such as image-text contrast~\cite{radford2021learning} as follows:
\begin{equation}
    \mathcal{L}_{\text{VLM}}=
    - \sum_{i=1}^N \log \frac{\exp{(z^{I}_{i}\cdot z^{T}_{i}/\tau)}}{\sum_{j=1}^{N}{\exp(z^I_i\cdot z^T_j/\tau)}} 
    - \sum_{i=1}^N \log \frac{\exp{(z^{T}_{i}\cdot z^{I}_{i}/\tau)}}{\sum_{j=1}^{N}{\exp(z^{T}_{i}\cdot z^{I}_{j}/\tau)}},
    \label{eq_vlm}
\end{equation}
where the two terms on the right denote image-to-text and text-to-image contrastive losses respectively. The notations $z^{I}_{i} = F^{I}(x^{I}_{i})$ and $z^{T}_{i} = F^{T}(x^{T}_{i})$ stand for the encoded image and text features respectively, $\tau$ denotes a temperature parameter~\cite{wu2018unsupervised}, and ``$\cdot$'' stands for the inner-product that measures the cosine similarity between two features.

\textbf{VLM inference.} A pre-trained VLM can perform open-vocabulary image recognition on arbitrary unlabelled target domains by reasoning on both images and texts~\cite{radford2021learning}. Given an arbitrary unlabelled target dataset $D = \{X^{I}, X^{T}\}$, $X^{I} = \{x^{I}_{n}\}_{n=1}^{N}$ stands for $N$ unlabelled images and $X^{T} = \{x^{T}_{m}\}_{m=1}^{M}$ denotes $M$ class names of interest, e.g., $X^{T} = \{\text{car, bus, ..., bike, person}\}$.
The pre-trained VLM predicts the probability of an image $x^{I}$ belonging to  class $x^{T}$ by:
\begin{equation}
    p_{x^{I} \rightarrow x^{T}}= z^{I} \cdot z^{T},
    \label{eq_vlm_infer}
\end{equation}
where $z^{I} = F^{I}(x^{I})$, $z^{T} = F^{T}(x^{T})$. Theoretically, VLMs can work with any class names $X^{T}$ and thus achieve open-vocabulary image recognition. Note $X^{T} = \{x^{T}_{m}\}_{m=1}^{M}$ contains $M$ target-domain class names but provides no information of which image belongs to which class name~\cite{radford2021learning}.

On the other hand, the performance of VLM is often constrained on target data ~\cite{zhou2022learning,li2022masked} due to cross-domain discrepancies in both image distributions and text distributions.

\subsection{
Definition of Open-vocabulary Domain Adaptation (OVDA)}
This work focuses on the task of OVDA, a new unsupervised domain adaptation (UDA) framework that transfers a pre-trained VLM $F = \{F^{I}, F^{T}\}$ towards an arbitrary unlabelled target domain $D = \{X^{I}, X^{T}\}$ with certain unsupervised training losses, i.e., $\mathcal{L}_{\text{OVDA}} = \mathcal{L}_{\text{unsupervised}}(X^{I}, X^{T}; F^{I}, F^{T})$. Take self-training~\cite{zhu2005semi,zou2018unsupervised} as an example. Given $X^{I} = \{x^{I}_{n}\}_{n=1}^{N}$ and $X^{T} = \{x^{T}_{m}\}_{m=1}^{M}$, the unsupervised training loss on unlabelled target data can be formulated as the following:

\begin{equation}
    y^{I}_{n} = \argmax_{m}  \ \ z^{I}_{n} \cdot z^{T}_{m}, \ \ \ \ \ \
    \mathcal{L}_{\text{ST}} =
    - \sum_{n=1}^N \log \frac{\sum_{m=1}^{M} \exp{(z^{I}_{n}\cdot z^{T}_{m}/\tau)} \times \mathbbm{1}(\hat{y}^{I}_{n} == m)}{\sum_{m=1}^{M}{\exp(z^{I}_{n}\cdot z^{T}_{m}/\tau)}},
    \label{eq_ovda_st}
\end{equation}
where $z^{I}_{n}$ and $z^{T}_{m}$ denote the encoded image and text features, i.e., $z^{I}_{n} = F^{I}(x^{I}_{n})$ and $z^{T}_{m} = F^{T}(x^{T}_{m})$. $y^{I}_{n}$ stands for the pseudo label of $x^{I}_{n}$. 

Note the unsupervised training is often unstable and susceptible to collapse if we optimize VLM image encoder and text encoder concurrently~\cite{li2022masked}. Hence, we freeze the VLM text encoder during unsupervised domain adaptation for stable adaptation. 

\subsection{Prompt Ensemble Self-training}

\begin{figure}[t]
\centering
\includegraphics[width=0.98\linewidth]{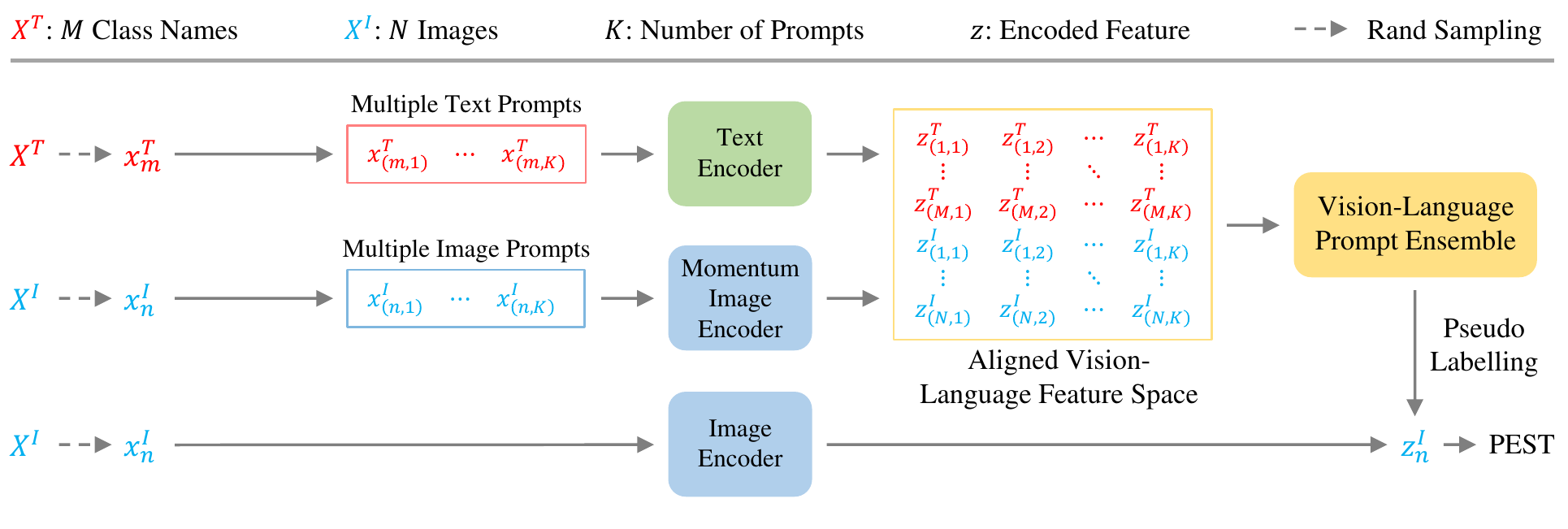}
\caption{
\textbf{Overview of Prompt Ensemble Self-training (PEST).} 
PEST exploits the complementary property of multiple prompts within and across vision and language modalities, which enables joint exploitation of vision and language information and effective learning of image-text correspondences in the unlabelled target domains. 
Besides, PEST captures temporal information via temporal prompt ensemble along the training process, which helps memorize previously learnt target information.
}
\label{maskedgan:ssm_bsm}
\end{figure}

We tackle the challenge of OVDA from a perspective of multi-prompt learning~\cite{jiang2020can,schick2020exploiting,yuan2021bartscore}.
As illustrated in Fig.~\ref{maskedgan:ssm_bsm}, we design prompt ensemble self-training (PEST) that introduces language prompt ensemble and vision prompt ensemble over self-training to mitigate the domain discrepancies in image and text distributions simultaneously.
In addition, PEST captures temporal information via temporal prompt ensemble, which helps memorize previously learnt target information by fusing the prompts encoded by the intermediate models evolved along the adaptation process.

\textbf{Language prompt ensemble} fuses the text features generated from different text prompts, aiming to leverage the complementary
information of multiple text prompts (i.e., various text descriptions for a class~\cite{luddecke2022image,zang2022open})
to mitigate the cross-domain discrepancy in text distributions.
It employs a Large Language Model~\cite{brown2020language,Ben2021Gpt-j-6b} (LLM) to generate multiple text prompts for a given class name and then encodes them by the VLM text encoder.
The encoded text features are then fused in a two-step manner: 1) uniformly average the multiple text features to acquire an initial prompt centroid 2) calculate the final prompt centroid by weighted average where the weight of each feature is the distance between it and the initial prompt centroid.
This two-step operation allows smooth prompt fusion by weighting down the effect of corner cases, which is important for language prompt ensemble as the LLM-generated prompts are not always reliable (e.g., when experiencing generation failures, LLM may generate only a full stop character ``.'' or a random word).

Given a class name $x^{T}_{m} \in X^{T}$, we employ the Large Language Model~\cite{brown2020language} to generate $K$ text prompts $\{x^{T}_{(m,1)}, x^{T}_{(m,2)}, ..., x^{T}_{(m,K)}\}$ and then the VLM text encoder $F^{T}$ to encode the generated prompts to acquire text features $\{z^{T}_{(m,1)}, z^{T}_{(m,2)}, ..., z^{T}_{(m,K)}\}$ (i.e., $z^{T}_{(m,k)} = F^{T}(x^{T}_{(m,k)})$).
The text features are then fused in a two-step manner to get the final text prompt centroid $\delta^{T}_{m}$:
\begin{equation}
    \delta^{T_{\text{initial}}}_{m} = \frac{1}{K} \sum_{k=1}^K z^{T}_{(m,k)}, \ \ \delta^{T}_{m} = \sum_{k=1}^K (z^{T}_{(m,k)} \cdot \delta^{T_{\text{initial}}}_{m}) \times z^{T}_{(m,k)},
    \label{eq_LPE}
\end{equation}
where ``$\cdot$'' denotes inner-product and $(z^{T}_{(m,k)} \cdot \delta^{T_{\text{initial}}}_{m})$ measures the distance between $z^{T}_{(m,k)}$ and $\delta^{T_{\text{initial}}}_{m}$.

\textbf{Vision prompt ensemble} fuses the image features generated from multiple image prompts, aiming to utilize the complementary property of multiple image prompts (i.e., various image descriptions for a class~\cite{luddecke2022image,zang2022open}) for mitigating the cross-domain discrepancy in image distributions.
Given an image, it employs certain off-the-shelf image augmentation policies~\cite{cubuk2020randaugment}
to generate multiple image prompts, encodes them with the VLM image encoder, and fuses the encoded image features in a class-wise manner.
Since target images are unlabelled, we generated pseudo labels for class-wise image feature fusion.
The class-wise feature fusion allows category-wise image prompt consolidation, which is crucial to visual prompt ensemble due to the abundance of target images and the encoded image features.
In addition, it simplifies vision-language prompt ensembling greatly (described in the later paragraphs) as language prompt ensemble also works in a category-wise manner.
Besides, with temporal prompt ensembling (described in the later paragraphs), it allows to dynamically select image prompts using pseudo labels along the adaptation process to describe each class visually.

Given an image $x^{I}_{n} \in X^{I}$, we adopt the off-the-shelf image augmentation policies in~\cite{cubuk2020randaugment} to generate $K$ image prompts $\{x^{I}_{(n,1)}, x^{I}_{(n,2)}, ..., x^{I}_{(n,K)}\}$ and then the VLM image encoder $F^{I}$ to encode the generated image prompts to acquire image features $\{z^{I}_{(n,1)}, z^{I}_{(n,2)}, ..., z^{I}_{(n,K)}\}$ (i.e., $z^{I}_{(n,k)} = F^{I}(x^{I}_{(n,k)})$).
Finally, the encoded features are fused in a class-wise manner to get the image prompt centroid $\delta^{I}_{m}$:
\begin{equation}
    \delta^{I}_{m} = \frac{1}{\sum_{n}^{N} \sum_{k=1}^K \mathbbm{1}(\hat{y}^{I}_{(n, k)} == m)} \sum_{n}^{N} \sum_{k=1}^K z^{I}_{(n, k)} \times \mathbbm{1}(\hat{y}^{I}_{(n, k)} == m), \ \ 
    \label{eq_VPE}
\end{equation}
where $\mathbbm{1}(\hat{y}^{I}_{(n, k)} == m)$ returns ``1'' if $\hat{y}^{I}_{(n, k)} = m$ else 0. Note $\hat{y}^{I}_{(n, k)} = \argmax_{m} z^{I}_{(n, k)} \cdot z^{T}_{m}$ denotes the pseudo label of $x^{I}_{(n, k)}$. Note we employ the momentum update of $F^{I}$ in the vision prompt ensemble for stable feature encoding and better capturing of temporal information, as shown in Fig.~\ref{maskedgan:ssm_bsm}.

\textbf{Temporal vision-language prompt ensemble} exploits the synergy between vision and language by fusing multiple text prompts and multiple image prompts over an aligned vision-language feature space.
It employs the text and image prompt centroids as starting point and updates them with the image prompt centroids generated by the intermediate VLM image encoder evolved along the adaptation process.
This enables prompt ensemble not only within but also across vision and language modalities, capturing the complementary advantages of vision and language information effectively.
In addition, the updating also achieves \textbf{temporal prompt ensemble} that captures previously learnt target information effectively.
Note we conduct temporal ensemble for image prompts only as the VLM text encoder is frozen during the adaptation process.

Specifically, we use the text and image prompt centroids $\delta^{T}_{m}$ and $\delta^{I}_{m}$ to initialize the image-text prompt centroid $\delta^{IT}_{m}$ and 
keep updating
$\delta^{IT}_{m}$ with $\delta^{I}_{m}$ 
along the adaptation process 
as follows:
\begin{equation}
    \delta^{IT_{\text{initial}}}_{m} = \delta^{I}_{m} + \delta^{T}_{m}, \ \ \ \ \delta^{IT*}_{m} \leftarrow \lambda \delta^{IT}_{m} + (1 - \lambda) \delta^{I}_{m},
    \label{eq_vlpe_update}
\end{equation}
where
$\delta^{IT}_{m}$ and $\delta^{IT*}_{m}$ denote the image-text prompt centroid before and after one update, respectively. $\lambda$ is a coefficient that controls the update speed of temporal ensemble. Note the first part denotes \textit{vision-language prompt ensemble} while the second part denotes \textit{temporal prompt ensemble}.

\textbf{Prompt ensemble self-training.}
Given image-text prompt centroid $\delta^{IT}_{m}$, target images, $X^{I} = \{x^{I}_{n}\}_{n=1}^{N}$ and target class names $X^{T} = \{x^{T}_{m}\}_{m=1}^{M}$, we employ $\delta^{IT}_{m}$ to “prompt” unsupervised self-training, which can be formulated as follows:
\begin{align}
    &y^{I}_{n} = \argmax_{m} \ \ (z^{I}_{n} \cdot z^{T}_{m}) \times (z^{I}_{n} \cdot \delta^{IT}_{m}),
    \\
    &\mathcal{L}_{\text{PEST}} =
    - \sum_{n=1}^N \log \frac{\sum_{m=1}^{M} \exp{(z^{I}_{n}\cdot z^{T}_{m}/\tau)} \times \mathbbm{1}(\hat{y}^{I}_{n} == m)}{\sum_{m=1}^{M}{\exp(z^{I}_{n}\cdot z^{T}_{m}/\tau)}},
    \label{eq_PEST}
\end{align}
where $z^{I}_{n}$ and $z^{T}_{m}$ denote the encoded image and text features, i.e., $z^{I}_{n} = F^{I}(x^{I}_{n})$ and $z^{T}_{m} = F^{T}(x^{T}_{m})$. $y^{I}_{n}$ stands for the pseudo label of $x^{I}_{n}$ generated with
$\delta^{IT}_{m}$.
The image-text prompt centroid $\delta^{IT}_{m}$ captures rich target image and text information. It is thus more invariant to visual and textual domain discrepancies and can “prompt” self-training to generate more accurate pseudo labels.

\section{Experiments}

\begin{table}[t]
\centering
\caption
{Image recognition datasets used for open-vocabulary domain adaptation benchmark.
}
\resizebox{\textwidth}{!}{
    \begin{tabular}	{l r r c l}
    \toprule
    Dataset  & Classes & Images & Domains & Description \\
    \midrule
    Office~\cite{saenko2010adapting} &31 &4,110 & 4  & Office objects from Amazon, DSLR, Webcam and Stnthetic domains.\\
    Office-home~\cite{venkateswara2017deep} &65 &15,588 & 4  & Office and Home objects from Art, Clipart, Product and Real-World domains.\\
    Adaptiope~\cite{ringwald2021adaptiope} &123 &36,900 &3  & Class-balanced object dataset with Product, Real-Life and Synthetic domains.\\
    VisDA~\cite{peng2017visda} & 12 &207,785 & 2 & A large-scale common object dataset with synthetic and real domains.\\ 
    \midrule
    ImageNet~\cite{deng2009imagenet} &1,000 &1,281,167 & 1   & A large-scale real-world object dataset with a wide range of categories. \\
    SUN397~\cite{xiao2010sun} & 397 & 76,129 &1   & A real-world indoor and outdoor scenes dataset for scene understanding. \\
    Food101~\cite{bossard2014food} &101 & 75,750 & 1  & A real-world food dish dataset for food recognition. \\
    GTSRB~\cite{stallkamp2011german} &43 & 26,640 &1 & A real-world german traffic sign dataset for sign recognition. \\
    DTD~\cite{cimpoi2014describing} &47 & 3,760 & 1   & A real-world describable texture image dataset for texture perception. \\
    UCF101~\cite{soomro2012ucf101} &101 & 9,537 & 1   & A real-world human action video dataset for action recognition.  \\
	\bottomrule
	\end{tabular}
}
\label{table:datasets}
\end{table}

\subsection{Datasets}

We benchmark our PEST extensively over 10 widely adopted image recognition datasets. As Table~\ref{table:datasets} shows, the 10 datasets have rich diversity, spanning multi-domain datasets with object images captured from several domains (e.g., real-world, synthetic, art, product and clipart domains) to single-domain datasets with real-world images for some specific visual task (e.g., the recognition of common objects, indoor and outdoor scenes, foods, traffic signs, natural textures and human actions).

\begin{table}[h]
\caption{OVDA performance on multi-domain datasets of Office, Office-Home and Adaptiope.
}
\begin{center}
\resizebox{\linewidth}{!}{
\begin{tabular}{lcccccccccccccc}
\toprule
\multirow{2}*{ViT-B/16} & \multicolumn{5}{c}{Office} & \multicolumn{5}{c}{Office-Home} & \multicolumn{4}{c}{Adaptiope
}\\
\cmidrule(lr){2-6} \cmidrule(lr){7-11} \cmidrule(lr){12-15}
&  \multicolumn{1}{c}{A}  & \multicolumn{1}{c}{W}   & \multicolumn{1}{c}{D} &  \multicolumn{1}{c}{S} &\multicolumn{1}{c}{Mean} & \multicolumn{1}{c}{A}   & \multicolumn{1}{c}{C} &\multicolumn{1}{c}{P} &\multicolumn{1}{c}{R} &\multicolumn{1}{c}{Mean} & \multicolumn{1}{c}{P} &\multicolumn{1}{c}{R} &\multicolumn{1}{c}{S} &\multicolumn{1}{c}{Mean} \\
\midrule
CLIP~\cite{radford2021learning} & 77.9&79.4&76.9&56.7&72.7&74.4&58.5&79.6&79.4&72.9&82.6&78.2&45.9&68.9
 \\
 ST~\cite{zhu2005semi} & 78.6&81.1&78.3&68.6&76.6&77.8&62.5&81.3&80.3&75.4&86.7	&82.0 &	49.5	&72.7\\
 CBST~\cite{zou2018unsupervised} & 79.1&80.7&78.5&68.9&76.8&77.3&62.8&81.7&80.7&75.6&86.9&83.2&50.1&73.4\\
 CRST~\cite{zou2019confidence} & 78.8&81.2&79.1&69.0&77.0&78.1&63.1&81.4&81.1&75.9&87.1&83.9&50.7&73.9\\
 SHOT~\cite{liang2020we} & 79.2&81.1&81.2&67.1&77.1&77.9&64.3&80.9&81.5&76.1&88.3&84.7&51.2&74.7\\
MUST~\cite{li2022masked} &  79.0&81.4&79.5&69.2&77.2&77.7&63.9&82.1&81.4&76.2&88.8&85.3&51.5&75.2
\\
\rowcolor{gray!16} PEST (Ours) & 84.3&82.8&81.3&72.3 & 80.1& 78.9&68.9&85.7&82.4&78.9 & 91.8&88.1&59.8&79.9
\\
\bottomrule
\toprule
\multirow{2}*{ResNet-50} & \multicolumn{5}{c}{Office} & \multicolumn{5}{c}{Office-Home} & \multicolumn{4}{c}{Adaptiope
}\\
\cmidrule(lr){2-6} \cmidrule(lr){7-11} \cmidrule(lr){12-15}
&  \multicolumn{1}{c}{A}  & \multicolumn{1}{c}{W}   & \multicolumn{1}{c}{D} &  \multicolumn{1}{c}{S} &\multicolumn{1}{c}{Mean} & \multicolumn{1}{c}{A}   & \multicolumn{1}{c}{C} &\multicolumn{1}{c}{P} &\multicolumn{1}{c}{R} &\multicolumn{1}{c}{Mean} & \multicolumn{1}{c}{P} &\multicolumn{1}{c}{R} &\multicolumn{1}{c}{S} &\multicolumn{1}{c}{Mean} \\
\midrule
CLIP~\cite{radford2021learning} & 72.9 &68.9&73.1&48.2& 65.7&64.6&42.1&71.9&71.9&62.6&74.5&66.2&35.8&58.8
 \\
 ST~\cite{zhu2005semi} & 75.2&66.8&71.3&44.1&64.3&66.7	&38.6&72.0	&73.8&62.7&75.7&70.7&26.7&57.7\\
 CBST~\cite{zou2018unsupervised} &  75.2 &67.8&72.2&51.1&66.5&68.1&	41.5&73.6&74.5&	64.4&77.2&71.1&34.3&60.8\\
 CRST~\cite{zou2019confidence} & 76.4&67.4&74.5&52.3&67.6&68.3&42.3&74.8&75.3&65.1&78.3&71.2&36.2&61.9\\
 SHOT~\cite{liang2020we} & 77.5&70.1&76.8&54.8&69.8&68.4&44.2&75.7&75.6&65.9&78.5&72.4&36.8&62.5\\
\rowcolor{gray!16} PEST (Ours) &79.6&75.3&80.3&55.0&72.5&68.6&47.9&78.2&77.4&68.0&80.7&75.6&37.8&64.7
\\
\bottomrule
\toprule
\multirow{2}*{ResNet-101} & \multicolumn{5}{c}{Office} & \multicolumn{5}{c}{Office-Home} & \multicolumn{4}{c}{Adaptiope
}\\
\cmidrule(lr){2-6} \cmidrule(lr){7-11} \cmidrule(lr){12-15}
&  \multicolumn{1}{c}{A}  & \multicolumn{1}{c}{W}   & \multicolumn{1}{c}{D} &  \multicolumn{1}{c}{S} &\multicolumn{1}{c}{Mean} & \multicolumn{1}{c}{A}   & \multicolumn{1}{c}{C} &\multicolumn{1}{c}{P} &\multicolumn{1}{c}{R} &\multicolumn{1}{c}{Mean} & \multicolumn{1}{c}{P} &\multicolumn{1}{c}{R} &\multicolumn{1}{c}{S} &\multicolumn{1}{c}{Mean} \\
\midrule
CLIP~\cite{radford2021learning} & 73.2&73.8&75.1&50.2&68.0&69.5&47.8&74.3&74.2&66.4&75.9&69.0&35.3&60.0
 \\
 ST~\cite{zhu2005semi} & 74.4&74.2&73.8&54.3&69.1&71.4&43.2&74.9&75.0&66.1&78.4&71.8&37.8&62.6\\
 CBST~\cite{zou2018unsupervised} & 74.6&75.9&72.9&58.1&70.3&72.3&44.9&77.7&76.2&67.7&79.5&73.3&41.5&64.7\\
 CRST~\cite{zou2019confidence} &75.3&76.6&73.4&58.5&70.9&73.4&45.9&78.4&76.8&68.6&80.1&75.2&43.7&66.3 \\
 SHOT~\cite{liang2020we} & 76.9&78.2&75.1&59.0&72.3&73.5&47.2&79.1&77.4&69.3&81.9&76.3&44.1&67.4\\
\rowcolor{gray!16} PEST (Ours) & 80.1&81.2&77.5&61.9&75.1&74.6&51.2&82.6&78.9&71.8&85.3&78.8&45.7&69.9
\\
\bottomrule
\end{tabular}
}
\end{center}
\label{tab:Office}
\end{table}

\subsection{Implementation Details}

We conduct experiments with three popular backbones, i.e., ResNet-50~\cite{he2016deep}, ResNet-101~\cite{he2016deep} and ViT-B~\cite{dosovitskiy2020image} pre-trained by CLIP~\cite{radford2021learning}. In training, we employ AdamW optimizer~\cite{loshchilov2017decoupled} with a weight decay of $0.05$, and set the initial learning rate as $1e-5$ which is
adjusted with a cosine learning rate schedule.
We use $2$ GPUs with batch size $64$ and the unsupervised adaptation training adds only a small amount of computation overhead after VLM pre-training.
We set input image size as $224 \times 224$ and employ data augmentation policies of ``RandomResizedCrop+Flip+RandAug''~\cite{cubuk2020randaugment} to generate multiple image prompts. The momentum VLM image encoder is updated with a momentum coefficient of $0.99$.
All results except on ImageNet are obtained with above implementation details.
For the large-scale ImageNet, we follow the implementations in~\cite{li2022masked}
and use 16 GPUs with batch size 1024.
During evaluation, we simply use the center-cropped image.

\subsection{OVDA on Multi-domain Datasets}
Tables~\ref{tab:Office}-\ref{table:visda} report the image classification results on 4 representative multi-domain datasets.
The experiments were
conducted with 3 representative backbones, i.e., ResNet-50, ResNet-101 and ViT-B/16.
It
can be seen that our PEST achieves superior performance consistently over various domains as compared with state-of-the-art methods.
Besides, PEST
outperforms CLIP substantially on Office (S)ynthetic domain, Office-Home (C)lipart domain and Adaptiope (S)ynthetic domain with 15.6\%, 10.4\% and 13.9\% accuracy improvement, respectively, showing that PEST can well handle the target domains with large domain discrepancies, i.e., Synthetic and Clipart styles.

\begin{table}[h]
\centering
\caption{
OVDA performance on large-scale multi-domain dataset VisDA.
}
\label{table:visda}
\resizebox{\textwidth}{!}{
\begin{tabular}{lccccccccccccc}
\toprule
\multicolumn{14}{c}{\textbf{VisDA Synthesis Domain}} \\\midrule
ViT-B/16 & plane & bcycl & bus & car & horse & knife & mcycl & person & plant & sktbrd & train & truck & Per-class \\
\midrule
CLIP~\cite{radford2021learning} &98.5&99.7&64.6&92.5&99.7&96.8&85.3&98.4&99.8&79.4&66.4&73.4&87.8  \\
ST~\cite{zhu2005semi} &97.2&99.9&60.4&84.5&99.8&98.6&92.5&99.7&99.9&79.3&74.2&84.4&89.2  \\
CBST~\cite{zou2018unsupervised} &98.4&99.7&67.3&85.2&99.8&99.1&95.3&99.9&99.4&83.4&83.4	&87.4	&91.5  \\
CRST~\cite{zou2019confidence}&98.1&98.2&70.5&86.5&98.6&98.7&94.3&98.8&97.8&86.7&88.7&86.1&91.9 \\
SHOT~\cite{liang2020we} &99.6&99.1&74.6&86.3&98.3&99.3&96.4&96.1&99.7&87.5&90.1&87.3&92.2  \\
MUST~\cite{li2022masked} &98.7&99.2&76.3&86.4&99.6&99.2&95.3&99.3&99.8&89.2&89.9&82.6&92.9\\
\rowcolor{gray!16} PEST (Ours)& 99.7&99.7&78.9&86.6&99.9&99.3&96.4&99.4&99.8&91.9&90.8&93.2&94.6 \\
\bottomrule
\toprule
\multicolumn{14}{c}{\textbf{VisDA Real Domain}} \\\midrule
ViT-B/16 & plane & bcycl & bus & car & horse & knife & mcycl & person & plant & sktbrd & train & truck & Per-class \\
\midrule
CLIP~\cite{radford2021learning}& 98.9&91.0&90.5&65.7&98.6&89.1&95.3&56.5&90.2&96.8&93.8&75.8&86.8 \\
ST~\cite{zhu2005semi} &99.4&87.3&92.5&68.3&98.1&90.4&94.6&69.3&91.2&96.7&94.5&66.4&87.3\\
CBST~\cite{zou2018unsupervised}&99.3&89.2&91.3&76.9&98.2&89.5&95.4&68.1&88.4&96.4&94.1&64.2&87.5 \\
CRST~\cite{zou2019confidence}&99.1&90.7&91.4&64.5&99.1&93.4&95.1&68.2&91.3&96.8&95.3&67.2&87.6 \\
SHOT~\cite{liang2020we} &99.3&92.8&91.9&65.3&98.7&95.2&94.5&67.7&92.1&96.9&95.4&67.9&88.1 \\
MUST~\cite{li2022masked} &99.2&95.7&92.6&56.9&99.1&98.6&96.0&67.0&	93.5&98.8&96.9&68.1	&88.5\\
\rowcolor{gray!16} PEST (Ours)& 99.2&95.9&92.1&66.1&99.2&97.8&96.7&70.8&92.7&98.4&96.2&74.6&90.0 \\
\bottomrule
\end{tabular}
}
\end{table}

\subsection{OVDA on Single-domain Datasets}
Table~\ref{table:single} reports the image classification over 5 popular single-domain datasets.
The experiments were conducted with 3 representative backbones, i.e., ResNet-50, ResNet-101 and ViT-B/16 (the results with ResNet-101 are provided in appendix B).
We can observe that PEST outperforms the state-of-the-arts by large margins consistently over different task-specific datasets, demonstrating that it can effectively handle various new visual recognition tasks by using unlabelled data. In addition, PEST brings substantial improvements upon CLIP over SUN397 (e.g., +11.0\% on ViT-B/16)
and GTSRB (e.g., +16.8\% on ViT-B/16),
showing that PEST can well tackle new image classification tasks with very specific objectives, e.g., indoor/outdoor scene and German traffic sign recognition.

\begin{table}[h]
\centering
\caption{
OVDA performance on single-domain datasets of various 
image recognition tasks.
}
\label{table:single}
\resizebox{\textwidth}{!}{
\begin{tabular}{lcccccccccccc}
\toprule
\multirow{2}*{Method} &\multicolumn{6}{c}{\textbf{ViT-B}} &\multicolumn{6}{c}{\textbf{ResNet-50}}\\\cmidrule(lr){2-7} \cmidrule(lr){8-13}
& SUN397 & Food101 & GTSRB & DTD & UCF101 &Mean & SUN397 & Food101 & GTSRB & DTD & UCF101 &Mean \\
\midrule
CLIP~\cite{radford2021learning} &60.8&85.6&32.5&44.5&64.1&57.5&54.0&73.1&25.0&39.8&56.0&49.5 \\
ST~\cite{zhu2005semi}&65.8&88.2&32.8&45.0&67.0&59.7& 59.0&74.4&20.5&35.8&56.4&49.2 \\
CBST~\cite{zou2018unsupervised}&63.2&89.5&37.6&44.3&68.1&60.5& 63.7&78.2&27.4&38.7&59.5&53.5  \\
CRST~\cite{zou2019confidence} &64.7&89.1&39.7&45.3&68.6&61.4&64.2&76.5&30.1&39.4&61.3&54.3\\
SHOT~\cite{liang2020we}&66.1&89.6&41.2&46.3&69.4&62.5&65.1&77.3&34.6&41.2&62.7&56.1 \\
MUST~\cite{li2022masked}&67.7&89.4&42.7&46.5&70.6&63.3& -&-&-&-&-&-\\
\rowcolor{gray!16} PEST (Ours)&71.8&91.1&49.3&52.7&73.9&67.7&65.7&79.5&39.6&49.4&65.6&59.9 \\
\bottomrule
\end{tabular}}
\end{table}

\subsection{OVDA on General Dataset ImageNet}
Table~\ref{table:imagenet} presents the image classification results on ImageNet.
It can be seen that PEST achieves superior performance as compared with state-of-the-art unsupervised methods, demonstrating the effectiveness of PEST over the very diverse and large-scale ImageNet.
Besides, PEST surpasses 16-shot supervised methods by a clear margin (i.e., +7.2\%), validating its advantages as a unsupervised method to mitigate the image annotation constraint and facilitate deep network training while handling new visual recognition tasks.

\begin{table}[h]
\centering
\caption
{Comparison with few-shot supervised adaptation methods and unsupervised adaption methods on ImageNet. All methods use the same CLIP ViT-B/16 model.
}
\resizebox{\textwidth}{!}{
\begin{tabular}	{l c c c c c c}
\toprule
\multirow{2}*{Method} & \multirow{2}*{CLIP~\cite{radford2021learning}} &\multicolumn{2}{c}{Supervised with 16 Labels per Class} & \multicolumn{3}{c}{Unsupervised}\\
\cmidrule(lr){3-4}\cmidrule(lr){5-7}
&  &CoCoOp~\cite{zhou2022conditional} &CoOp~\cite{zhou2022learning} &ST~\cite{zhu2005semi} &MUST~\cite{li2022masked} & \cellcolor{gray!16}{PEST (Ours)} \\\midrule
ImageNet Accuracy &  68.3 &71.0 & 71.5 &76.5 & 77.7 &\cellcolor{gray!16}{78.7}\\
\bottomrule
\end{tabular}
}
\label{table:imagenet}
\end{table}

\subsection{Discussion}

\textbf{Generalization across different domains and tasks.} We examine the generalization of
PEST with respect to image recognition tasks and domains.
Specifically, we perform extensive evaluations over 10 widely studied multi-domain~\cite{saenko2010adapting,venkateswara2017deep,ringwald2021adaptiope,peng2017visda} and single-domain~\cite{deng2009imagenet,xiao2010sun,bossard2014food,stallkamp2011german,cimpoi2014describing,soomro2012ucf101} datasets as described in Table~\ref{table:datasets}.
Experimental results in Tables~\ref{tab:Office}-~\ref{table:imagenet} show that the proposed PEST achieves superior image recognition performance consistently
across different domains and tasks.

\textbf{Generalization across different backbones.} We study the generalization of PEST by assessing it with three popular image recognition backbones,
including two CNNs (i.e., ResNet-50 and ResNet-101) and one Transformer (i.e., ViT-B/16).
Results in Tables~\ref{tab:Office}-~\ref{table:imagenet} and the tables in appendix B show that our PEST works effectively and consistently over different image recognition backbones.

\begin{table}[h]
\caption{
Ablation studies of PEST with ViT-B/16 on Office dataset.
}
\begin{center}
\resizebox{0.98\linewidth}{!}{
\begin{tabular}{lccccc}
\toprule
\multirow{2}*{Method} & \multicolumn{2}{c}{Vision-Language Prompt Ensemble} & \multirow{2}*{Temporal Prompt Ensemble} & \multirow{2}*{Office (Mean)}\\
\cmidrule(lr){2-3}
&  \multicolumn{1}{c}{Vision Prompt Ensemble}  & \multicolumn{1}{c}{Language Prompt Ensemble}  \\
\midrule
CLIP~\cite{radford2021learning} &&& &72.7\\
ST~\cite{zhu2005semi} &&&&76.6\\
\midrule
&\checkmark&&&77.5\\
&&\checkmark&&78.2\\
&\checkmark&\checkmark&&78.7\\
\rowcolor{gray!16} PEST &\checkmark&\checkmark&\checkmark&{80.1}\\\bottomrule
\end{tabular}
}
\end{center}
\vspace{-1mm}
\label{table:ablation}
\end{table}

\textbf{Ablation study.} We conduct ablation studies with ViT-B/16 on Office as shown in Table~\ref{table:ablation}. As the core of the proposed PEST, we examine how our designed \textit{vision prompt ensemble}, \textit{language prompt ensemble} and \textit{temporal prompt ensemble} contribute to the overall performance of open-vocabulary domain adaptation. 
As shown in Table~\ref{table:ablation}, including 
either vision prompt ensemble or language prompt ensemble above self-training improves performance clearly,
showing that image and text prompts ensembling
help mitigate cross-domain discrepancies in image distributions and text distributions and can ``prompt'' unsupervised self-training with more accurate pseudo label prediction. 
In addition, combining vision and language prompt ensemble performs clearly better, indicating that the two types of prompt ensembling complement each other by working from orthogonal vision and language perspectives.
Furthermore, including \textit{temporal prompt ensemble} upon vision-language prompt ensemble (PEST in the last row) performs the best. It demonstrates the importance of temporal prompt ensemble that helps memorize previously learnt target information along the training process.

\begin{wraptable}{r}{5.8cm}
\vspace{-3ex}
\caption{
\small	
Parameter ablations with ViT-B/16 on Office. The default is marked in \colorbox{gray!16}{gray}.}
\small
\setlength\tabcolsep{3pt}
\resizebox{5.8cm}{!}{
\centering	
\begin{tabular}	{l | cccc } 
Parameter $\lambda$ & 0.9 &\cellcolor{gray!16}{0.99} &0.999 &0.9999\\
\midrule
Office (Mean) & 79.6 & \cellcolor{gray!16}{80.1} & 80.1 & 80.0 \\
\bottomrule
\end{tabular}
}
\vspace{-1ex}
\label{table:para}
\end{wraptable}	

\textbf{Parameter study.}
The parameter $\lambda$ in Eq.~\ref{eq_vlpe_update} controls the update speed of temporal ensemble. We investigate $\lambda$ by varying it from $0.9$ to $0.9999$ progressively, as shown in Table~\ref{table:para}.
It can be seen that varying $\lambda$ does not affect OVDA clearly.
The performance drops a bit while $\lambda = 0.9$, largely because a fast update may lead to unstable temporal prompt ensemble that only captures local information within each training batch.

\begin{wraptable}{r}{5.8cm}
\vspace{-0.5ex}
\caption{
\small	
Comparison with other multi-prompt learning methods with ViT-B/16 on Office.}
\small
\setlength\tabcolsep{3pt}
\resizebox{5.8cm}{!}{
\centering	
\begin{tabular}	{l | c} 
Method & Office (Mean)\\
\midrule
ST + Uniform Averaging~\cite{jiang2020can} & 77.2\\
ST + Weighted Averaging~\cite{qin2021learning} & 77.4\\
ST + Majority Voting~\cite{lester2021power} & 77.1 \\
\rowcolor{gray!16} PEST (Ours) & 80.1\\
\bottomrule
\end{tabular}
}
\vspace{-2ex}
\label{table:mpl}
\end{wraptable}

\textbf{Comparison with multi-prompt learning methods.}
We compare PEST with multi-prompt learning strategies that explore complementary advantages of different prompts via uniform averaging~\cite{jiang2020can,schick2020exploiting,yuan2021bartscore}, weighted averaging~\cite{jiang2020can,qin2021learning,schick2020exploiting}, majority voting~\cite{lester2021power,hambardzumyan2021warp}.
As Table~\ref{table:mpl} shows, existing multi-prompt learning methods do not perform well, largely because they were designed for NLP without considering the joint exploitation of vision and language modalities and the information memorization during unsupervised transfer. PEST instead learns and memorizes effective image-text correspondences in the unlabelled target domains via joint exploitation of vision and language information, which are essential to OVDA.

\begin{figure}[t]
\centering
\includegraphics[width=0.24\linewidth]{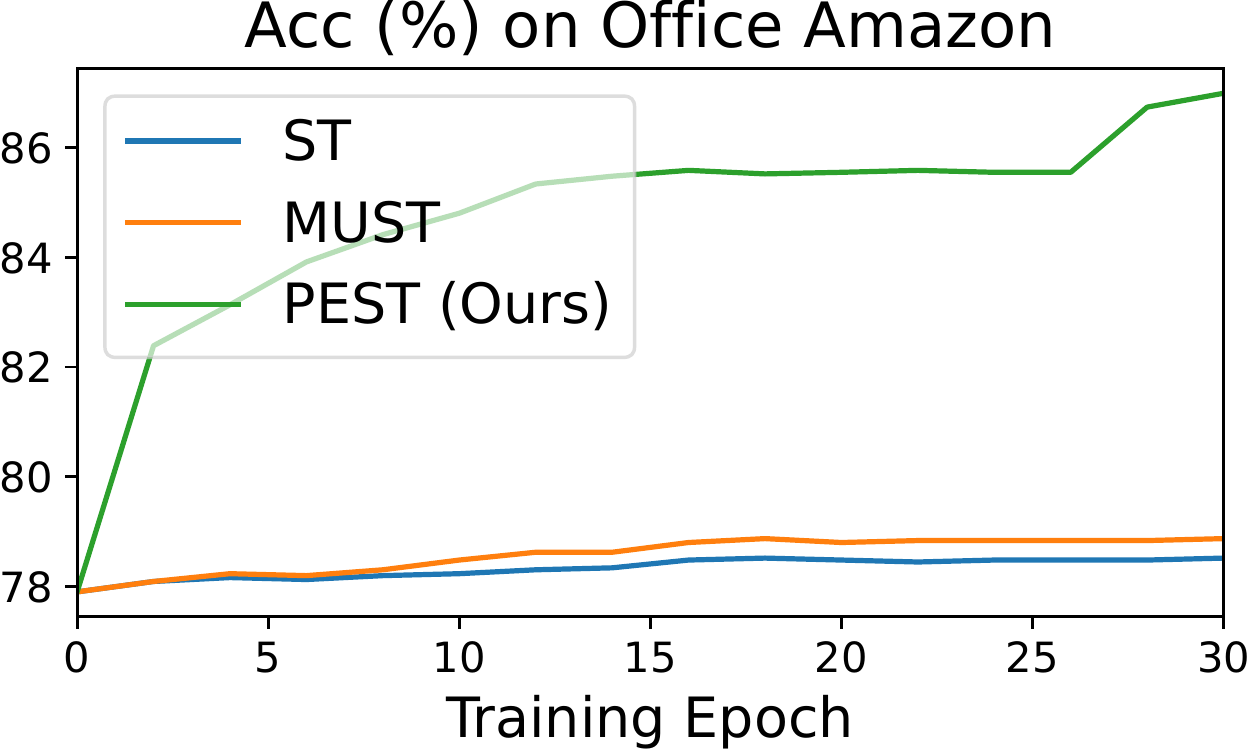}
\includegraphics[width=0.24\linewidth]{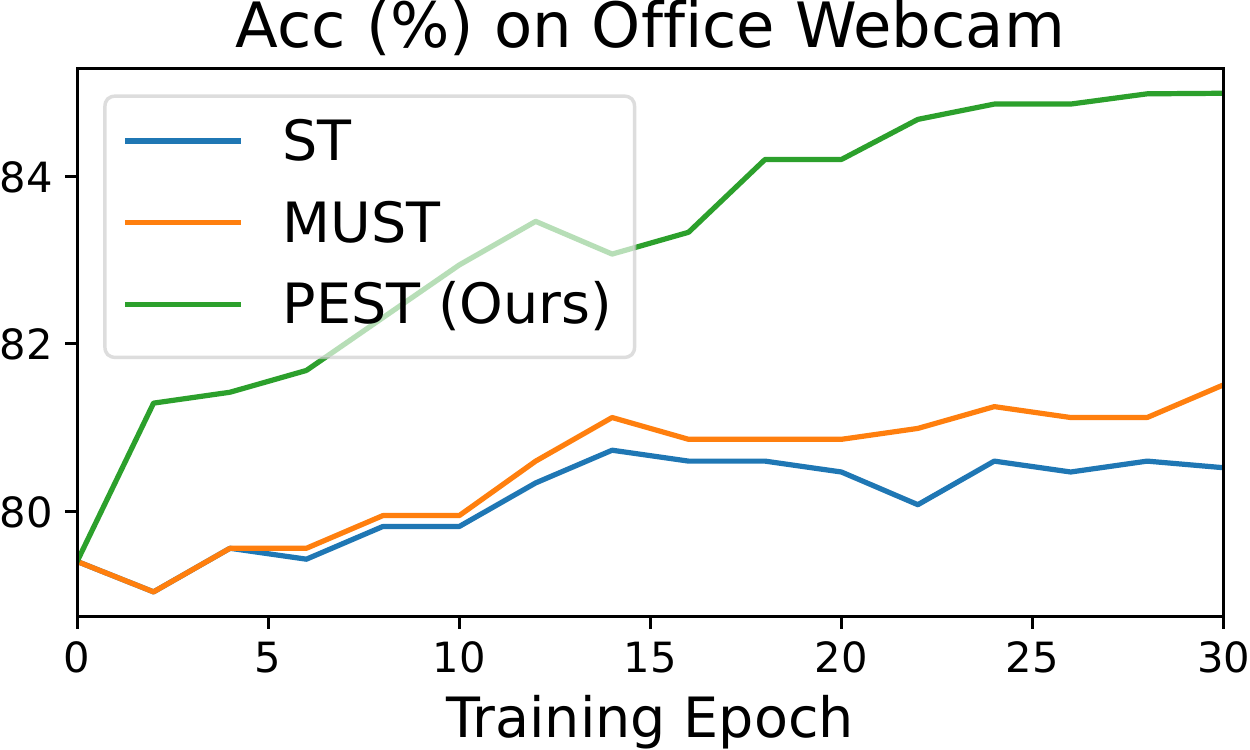}
\includegraphics[width=0.24\linewidth]{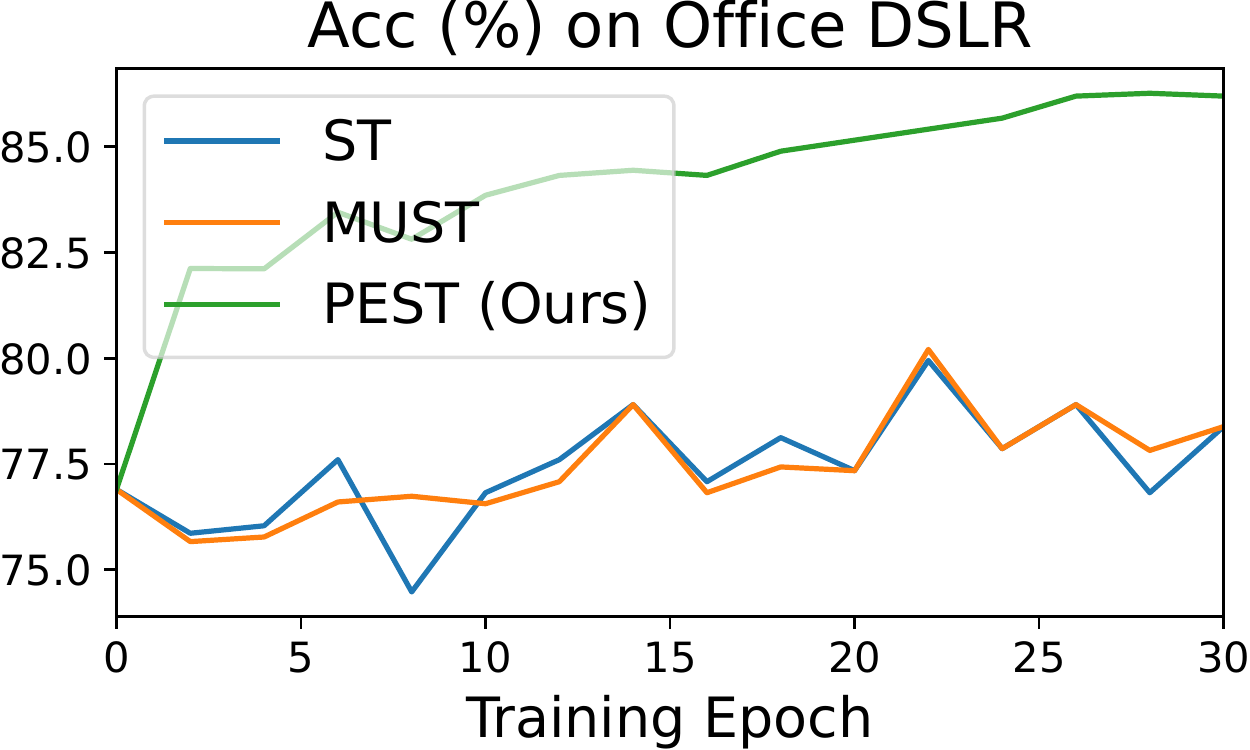}
\includegraphics[width=0.24\linewidth]{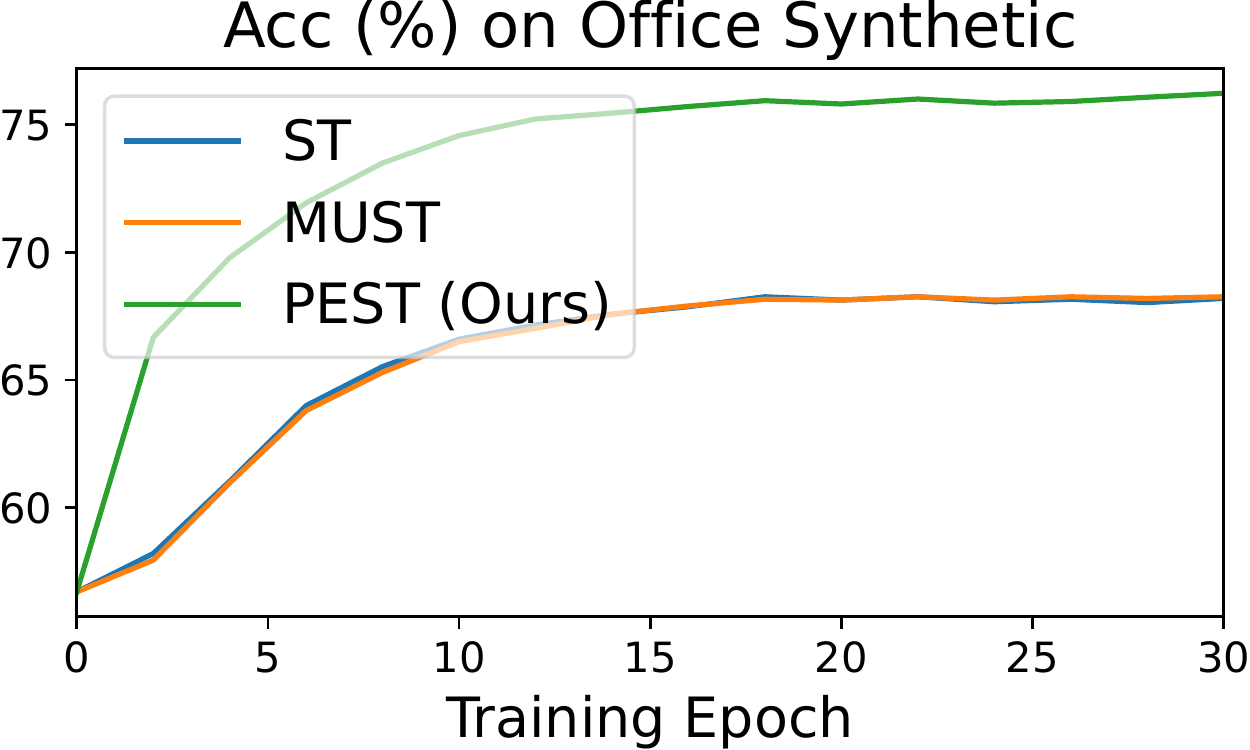}
\\
\caption{Pseudo label accuracy along the unsupervised adaptation process in OVDA (with ViT-B/16).
}
\vspace{-2ex}
\label{fig:Pseudo}
\end{figure}

\textbf{Pseudo label accuracy.}
Fig.~\ref{fig:Pseudo} shows the pseudo label accuracy along the unsupervised adaptation process. PEST generates much more accurate pseudo labels than the vanilla self-training (ST) and the state-of-the-art MUST. The superior pseudo label accuracy is largely attributed to the proposed prompt ensemble which helps capture rich target image and text information that is more invariant to visual and textual domain discrepancies and can “prompt” better unsupervised self-training.

\textbf{Comparison with MUST.}
MUST~\cite{li2022masked} tackles unsupervised adaptation of VLMs from the perspective of Masked Image Modelling~\cite{he2021masked} that heavily relies on Transformer backbones~\cite{dosovitskiy2020image}.
As a comparison, the proposed PEST works from the perspective of multiple-prompt learning that is independent to vision backbones.
Thus, PEST can seamlessly work on different vision backbones like CNNs and Transformers as shown in Tables~\ref{tab:Office}-\ref{table:single}.
In addition, Tables~\ref{tab:Office}-\ref{table:imagenet} show that PEST outperforms MUST clearly, largely because 
MUST exploits vision information largely while PEST exploits both vision and language information jointly which is better aligned with the objective of OVDA.

\begin{wraptable}{r}{5.8cm}
\vspace{-3ex}
\caption{
\small	
Results of w/ and w/o prompt engineering with ViT-B/16 on 5 tasks of SUN397, Food101, GTSRB, DTD and UCF101.}
\small
\setlength\tabcolsep{3pt}
\resizebox{5.8cm}{!}{
\centering	
\begin{tabular}	{l | c} 
Method & 5-task Mean\\
\midrule
CLIP w/o Prompt Engineering & 57.5\\
CLIP w/ Prompt Engineering & 62.0\\
MUST w/o Prompt Engineering & 63.3 \\
MUST w/ Prompt Engineering & 65.8 \\
\rowcolor{gray!16} PEST w/o Prompt Engineering & 67.7\\
\bottomrule
\end{tabular}
}
\vspace{-1ex}
\label{table:prompt_eng}
\end{wraptable}

\textbf{Prompt engineering.}
Both CLIP~\cite{radford2021learning} and MUST~\cite{li2022masked} mitigate the cross-domain text distribution gap by prompt engineering~\cite{radford2021learning} and ensembling, e.g., uniform averaging of 80 hand-crafted prompt templates
on ImageNet. Although hand-crafted prompt templates bring clear gains, 
manually designing prompts for each new image recognition task and domain is laborious and time-consuming and degrades the scalability greatly. 
As Table~\ref{table:prompt_eng} shows, without any prompt engineering, the proposed PEST still outperforms CLIP and MUST with clear margins, demonstrating its effectiveness and efficiency in handling new visual recognition tasks without prompt engineering.
Note we did not include multi-domain datasets~\cite{saenko2010adapting,venkateswara2017deep,ringwald2021adaptiope,peng2017visda} in this experiment due to the lack of hand-crafted prompt templates.

Due to the space limit, we provide more dataset details, experiments and discussions in the appendix.

\section{Conclusion}

This paper presents PEST, a novel open-vocabulary domain adaptation framework that explores multi-prompt learning to learn effective image-text correspondences over unlabelled target images. PEST exploits prompt ensemble over vision, language and temporal dimensions for simultaneous mitigation of image and text discrepancies across domains. It requires merely a pre-trained VLM but achieves effective and efficient UDA towards arbitrary unlabelled target domains, demonstrating its superiority in facilitating deep network training while handling arbitrary new visual recognition tasks and domains. Extensive experiments show that PEST achieves superb recognition performance consistently across different backbones and image recognition tasks and domains.
Moving forward, we will explore multi-prompt learning for other vision tasks such image generation.

\bibliographystyle{plain}
\small\bibliography{egbib}

\end{document}